\documentclass[graybox]{svmult}
\usepackage[first=0,last=9]{lcg}
\usepackage{scalerel}   % to scale subscripts

\usepackage{wrapfig}
\usepackage{graphicx}
\DeclareGraphicsExtensions{.pdf,.png,.jpg}

\usepackage{mathptmx}       % selects Times Roman as basic font
\usepackage{helvet}         % selects Helvetica as sans-serif font
\usepackage{courier}        % selects Courier as typewriter font
\usepackage{makeidx}         % allows index generation
\usepackage{graphicx}        % standard LaTeX graphics tool
\usepackage{multicol}        % used for the two-column index
\usepackage{wrapfig}

\makeindex             % used for the subject index
\usepackage{amssymb}
\usepackage{amsmath}
\usepackage{algorithm}
\usepackage{algorithmic}
\usepackage{hyperref} % <— add this line

\usepackage{multirow}
\usepackage{array}
\usepackage{tikz}           % to graph neural nets
\usepackage{color, colortbl}
\usepackage{makecell}

\definecolor{Gray}{gray}{0.9}

\begin{document}

\title*{Integrating Product Coefficients for Improved 3D LiDAR Data Classification (Part II)}

\author{Patricia Medina, Rasika Karkare}

\institute{
Patricia  Medina \at New York City College of Technology, CUNY; \email{patriciamg90@gmail.com}\\
Rasika Karkare \at Worcester Polytechnic Institute; \email{rskarkare@wpi.edu}}

\maketitle

%\abstract*{Each chapter should be preceded by an abstract (10--15 lines long) that summarizes the content. The abstract will appear \textit{online} at \url{www.SpringerLink.com} and be available with unrestricted access. This allows unregistered users to read the abstract as a teaser for the complete chapter. As a general rule the abstracts will not appear in the printed version of your book unless it is the style of your particular book or that of the series to which your book belongs.
%Please use the 'starred' version of the new Springer \texttt{abstract} command for typesetting the text of the online abstracts (cf. source file of this chapter template \texttt{abstract}) and include them with the source files of your manuscript. Use the plain \texttt{abstract} command if the abstract is also to appear in the printed version of the book.}

\abstract{This work extends our previous study on enhancing 3D LiDAR point-cloud classification with product coefficients \cite{medina2025integratingproductcoefficientsimproved}, measure-theoretic descriptors that complement the original spatial Lidar features. Here, we show that combining product coefficients with an autoencoder representation and a KNN classifier delivers consistent performance gains over both PCA-based baselines and our earlier framework. We also investigate the effect of adding product coefficients level by level, revealing a clear trend: richer sets of coefficients systematically improve class separability and overall accuracy. The results highlight the value of combining hierarchical product-coefficient features with autoencoders to push LiDAR classification performance further.}

\section{Introduction}
LiDAR point clouds, representing detailed three-dimensional descriptions of natural and built environments, are widely used in applications such as updating digital elevation models, monitoring glaciers and landslides, shoreline analysis, and urban development. A crucial step in these applications is the classification of 3D LiDAR points into semantic categories such as vegetation, man-made structures, and water.

In our previous work \cite{medina2025integratingproductcoefficientsimproved}, we introduced product coefficients as measure-theoretic descriptors that enrich LiDAR data with local structural information. Computed on dyadic neighborhoods around each point, these coefficients capture geometric variability beyond raw spatial coordinates. By augmenting the original features with these coefficients and applying Principal Component Analysis (PCA) for dimensionality reduction, we showed that standard classifiers such as k-nearest neighbors (KNN) and Random Forest achieved substantial improvements in accuracy and F1-macro scores compared to models trained only on spatial coordinates.

The present study extends that framework by introducing a nonlinear representation learning approach based on an autoencoder. Specifically, we replace PCA with a PyTorch-implemented autoencoder that learns compact latent representations directly from the enriched feature space consisting of the original features and the product coefficients. The autoencoder comprises two hidden layers with dimensions 30 and a latent dimension equal to the number of features. It was trained with a batch size of 256 for 30 epochs using the Adam optimizer and ReLU and LeakyReLU activation functions.

Experimental results demonstrate that the Autoencoder plus Product Coefficients framework consistently outperforms the previous PCA plus Product Coefficients approach, achieving higher classification accuracy and F1-macro scores. This improvement highlights the advantage of nonlinear representation learning in capturing complex feature dependencies and reducing redundancy more effectively than linear transformations such as PCA.

LiDAR (Light Detection and Ranging) technology itself remains at the core of this study. By transmitting laser pulses and measuring their return time, LiDAR systems generate precise 3D point clouds representing surface reflectivity and elevation \cite{inbook}. These datasets can be visualized and processed within GIS platforms such as ArcGIS \cite{arcgis} and are employed in various geoscientific and environmental applications including terrain mapping, forest inventory, shoreline detection, landslide risk analysis, and urban planning \cite{Moskal}.

Our focus remains on point-wise classification, assigning a class label to each point rather than to aggregated shapes. The proposed framework systematically integrates feature engineering through product coefficients, nonlinear dimensionality reduction using an autoencoder, and traditional classifiers such as KNN and Random Forest to assess performance improvements.

The paper is organized as follows. Section~\ref{sec:data} describes the LiDAR dataset and its attributes. Section~\ref{PCs} provides a brief review of product coefficients and their computation on dyadic sets. Section~\ref{subsec: PCs on lidar} details their computation on 3D LiDAR point clouds embedded in the unit cube. Section~\ref{subsec: PCA and PCs experients} introduces the autoencoder architecture, experimental design, and comparative results. Section~\ref{summary} concludes with a discussion of the findings and outlines future directions.

\section{Prior Work and Key Results from Our First Study}
In our previous work (see \cite{medina2025integratingproductcoefficientsimproved}), we introduced product coefficients as measure-theoretic features to enrich 3D LiDAR point-cloud classification. A lot pf the exposition on the background in needed on this paper is taken from \cite{medina2025integratingproductcoefficientsimproved}. Computed from local neighborhoods using a dyadic partitioning scheme, these coefficients were appended to spatial coordinates to provide classifiers with descriptors that capture structural information beyond raw geometry.

We evaluated this feature design by combining product coefficients with PCA and testing them with KNN and Random Forest classifiers. Three setups were compared: using only spatial coordinates, adding product coefficients directly, and applying PCA to the enriched feature set. The experiments showed that PCA was essential to unlocking the value of the coefficients, producing representations that classifiers could exploit more effectively.

The study demonstrated consistent improvements over the baseline, reaching accuracies of around 0.85 in the best configuration. These results confirmed the utility of product coefficients and pointed to the potential of pairing them with more flexible representation learning methods. That motivation forms the basis for the present work.

\section{The Data}\label{sec:data}

Lidar points can be classified into several categories, such as bare earth or ground, top of canopy, and water. These classes are represented by integer codes in the LAS files, as defined by the American Society for Photogrammetry and Remote Sensing (ASPRS) for LAS formats. The latest version includes eighteen classes: 0 for Never classified, 1 for Unassigned, 2 for Ground, 3 for Low vegetation, 4 for Medium vegetation, 5 for High vegetation, 6 for Building, 7 for Noise, 8 for Model key/Reserved, 9 for Water, 10 for Rail, 11 for Road surface, 17 for Bridge deck, and 18 for High noise.

In our experiments, we use a publicly available Lidar data used many years in one of the LAS tool
tutorials. The scene consists of a main tree from a Washington State park (mention credits to University of Washington remote sensing lab) contains 798452 data points. It has 3 classes class 1 - Ground, class 2 - Trunk/ branches and class 3 - canopy.\\
The number of data points in each class is: 

class 1 : 311208;
class 2 : 405911;
class 3 : 81333.

% 277, 572 points and includes five classes (ground, high vegetation, building and low vegetation.) A plot made in Python colored by class is shown in Fig. \,\ref{fig:scenario}. Note that the three classes of building, high vegetation and ground are the most visible in the this plot. 

\begin{figure}
	\centering	
	\includegraphics[scale=0.2]{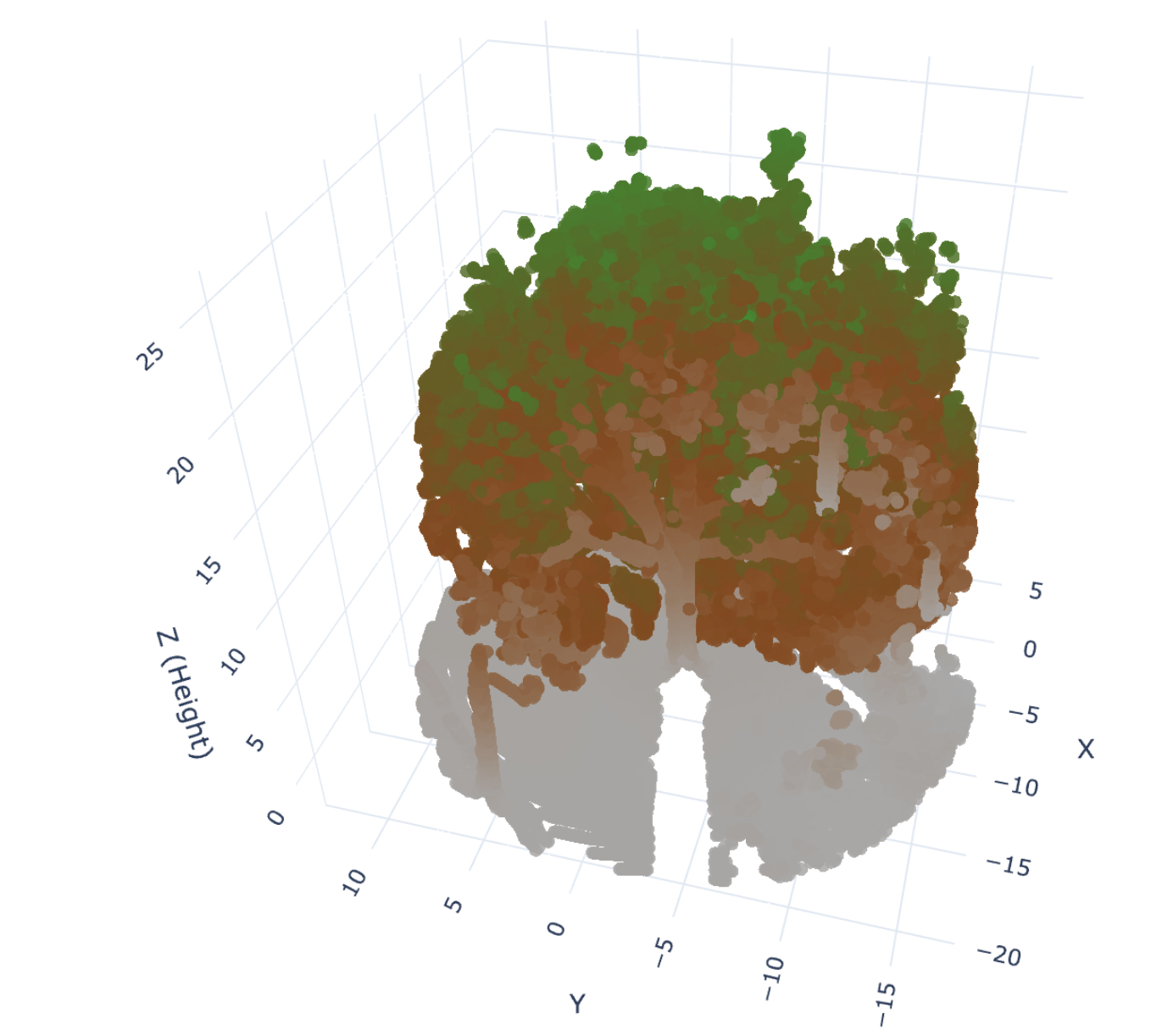}
	\caption{Plot representing LiDAR point cloud of a tree from a Washington state park. Exact coordinates are available. We consider only
		three classes (with the three more visible ones being ground, canopy and trunk). Data was provided by Prof. Monica Moskal, Remote Sensing and Geospatial Analysis Lab (RSGAL), University of Washington.}	
	\label{fig:scenario}
\end{figure}

\begin{figure}\label{fig:tree}
\centering	
\includegraphics[scale=0.8]{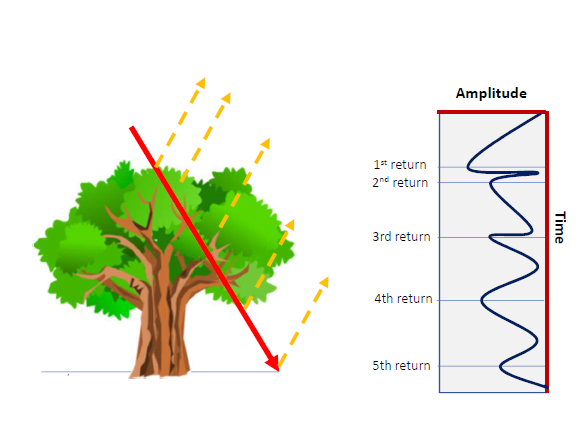}
\caption{\scriptsize A pulse can be reflected off a tree's trunk, branches, and foliage as well as reflected off the ground. This originates one of the data features (number of returns) This figure was used for the first time in one of the author's paper (see \cite{medina2019heuristic}) }
\end{figure}

Lidar point cloud attributes or features are typically: 
\begin{itemize}
	\item \textbf{Intensity}: The strength of the return signal, which measures the peak amplitude of pulses reflected back to the LiDAR detector. Higher intensity values often indicate reflective surfaces, aiding in material differentiation (e.g., vegetation vs. ground).
	
	\item \textbf{Return Number}: Specifies the order in which returns from a single emitted pulse are recorded. For example, a pulse hitting multiple objects may generate multiple returns: return number one for the first surface, return number two for the second, and so on. This attribute provides insight into the vertical structure of objects encountered by the pulse.
	
	\item \textbf{Number of Returns}: The total number of returns for a single emitted pulse, indicating how many reflective surfaces were encountered along the pulse path. Pulses passing through objects like vegetation can have multiple returns, while those hitting solid surfaces typically yield only one. See Fig.\,\ref{fig:tree} for an illustration of how the measures for this feature are collected.
	
	\item \textbf{Point Classification}: Defines the type of surface or object reflecting the pulse, with each category represented by an integer code (e.g., 2 for ground, 5 for vegetation canopy). Classifications are often added during post-processing and can help distinguish between terrain features and objects, such as buildings or water bodies.
	
	\item \textbf{Edge of Flight Line}: A binary flag (0 or 1) where 1 denotes points at the edge of the flight line and 0 indicates points within the main survey area. This helps identify data quality variations that can occur at flight edges.
	
	\item \textbf{RGB}: Red, Green, and Blue values often sourced from synchronized imagery taken with the LiDAR survey. This attribute enables enhanced visualization by adding color information to points, useful for applications such as vegetation mapping or urban analysis.
	
	\item \textbf{GPS Time}: The GPS timestamp (in seconds of the GPS week) marking the exact time the pulse was emitted. This attribute is critical for aligning LiDAR data with positional information, enabling accurate 3D reconstruction of the surveyed area.
	
	\item \textbf{Scan Angle}: The emission angle of the laser pulse, ranging from -90 to +90 degrees. A value of 0 degrees indicates a nadir shot (directly below the aircraft), while negative and positive values represent angles to the left and right, respectively. The scan angle aids in understanding the perspective and potential distortion in the captured data.
	
	\item \textbf{Scan Direction}: Indicates the movement direction of the laser scanning mirror when the pulse was emitted. A value of 1 means a positive direction (left to right in the flight path), and 0 means a negative direction (right to left). This attribute is useful for correcting any directional biases in the data.
\end{itemize}

For our experiments in this paper we only keep the spatial coordinates $(x,y,z)$ for the simplest classification and generate new features by computing product coefficients only using these three spatial coordinates from the original data.

%%%%%%%%%%%%%%%%

\section{Product coefficients}\label{PCs}
%\label{sec:methods}

In this section, we introduce the theoretical definition of product coefficients. We show how to compute product coefficients on a binary tree stemming from a set $X.$

Many datasets consist of streaming sets of vectors, which can be understood as (streaming) discrete metric-measure spaces. These datasets allow for the automatic computation of canonical multi-scale representations, guaranteed to exist based on mathematical theorems. These canonical representations are domain-agnostic, enabling them to be fused to describe what is  ``normal'' and, by extension, identify what is anomalous. Furthermore, these representations provide automatically generated canonical features that can be used in decision-making algorithms, such as those employed in machine learning.

We want to exploit the product coefficient parameters for inference and decision. 
Product coefficients can be derived from data samples and utilized to estimate unknown measures represented by those samples. For instance, given a collection of \( n \) samples of points from a dyadic set $X$ and a method to preprocess each sample into measures for the dyadic subsets (e.g., through counting measures), a corresponding set of product coefficients can be calculated for each sample. Since the product coefficient parameters uniquely distinguish measures determined by samples (after preprocessing the samples into measures) , they can be used as features for decision rules.

We focus on measures defined on dyadic sets which are sets with an ordered binary tree of
subsets. An example is the partition of the unit interval into dyadic subintervals. The measures are defined on the sigma algebra generated by the subsets in the binary tree.
The dyadic product formula representation lemma (lemma\, \ref{lemma1}) provides an explicit set of product coefficient parameters which are sufficient to distinguish measures on dyadic sets. Since LiDAR is geospatial and this type of data includes geometric aspects (such as those found in the ground and trees), product coefficients reflect these characteristics.

We compute the product coefficient quantities on a set $X$ normalized to fit the unit cube ${[0,1]}^3.$ The exposition and theory are based in the work by authors in \cite{Fefferman} and \cite{Linda}.

We first recall that a dyadic set is a collection of subsets structured as an ordered binary tree (e.g. unit interval, feature sets, unit cubes). More precisely, we consider a {\it dyadic} set $X$ 
which is the {\it parent} set or root of the ancestor tree of a system of left and right {\it child} subsets. For each subset $S$ (dyadic subset) of $X$, we denote the left child by $L(S)$ and the right child by $R(S).$  Let $\mu$ be a non-negative measure on $X$ and $dy$ the naive measure, such that $dy(X)=1$.  
\begin{equation*}
	dy(L(S))=\frac{1}{2}dy(S),\quad dy(R(S))=\frac{1}{2}dy(S)
\end{equation*}

Note that $\mu$ is additive in the binary set system, i.e. $\mu (S)= \mu(L(S) \cup R(S))=\mu(L(S))+ \mu(R(S))$ ($L(S)$ and $R(S)$ are disjoint.)

%%%  here is the general slide shown as a amotivation in original presentation %%%
\begin{figure}
\centering	
\includegraphics[trim={7cm 12cm 10cm 1cm},clip,scale=0.6]{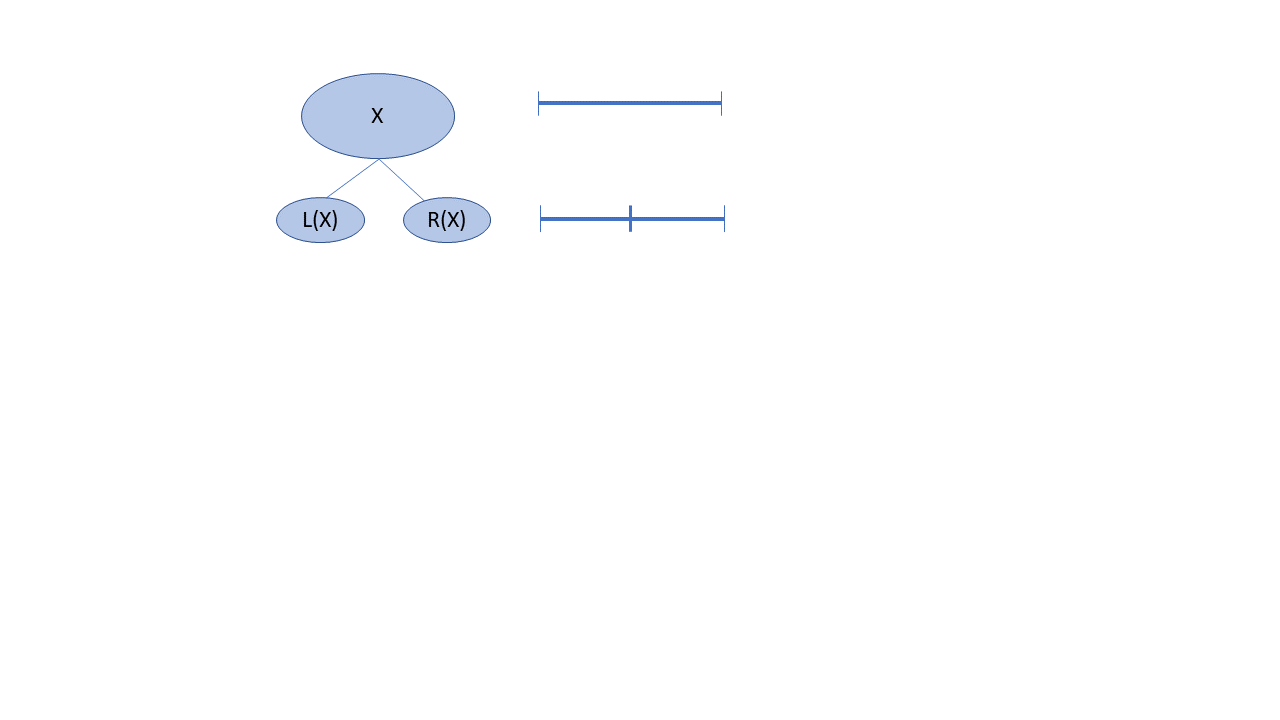}
\caption{illustration of the first level of a dyadic set or binary tree for a set $X$. On the right we have how this translate into an interval}
\label{fig:dyadic}
\end{figure}

%$X$: {\it parent} set or root of the ancestor tree of a system of left and right {\it child} subsets.

$\mu$: non-negative measure on $X$; 
On Fig.\,\ref{fig:dyadic}, we can consider $dy$ the naive measure, such that $dy(X)=1$ if we work for instance in the interval $[0,1].$
\begin{equation*}
	dy(L(S))=\frac{1}{2}dy(S),\quad dy(R(S))=\frac{1}{2}dy(S)
\end{equation*}
$\mu$ is additive in the binary set system,
 i.e. $\mu (S)= \mu(L(S) \cup R(S))=\mu(L(S))+ \mu(R(S))$ ($L(S)$ and $R(S)$ are disjoint.)

Let $\mu$ be a dyadic measure on a dyadic set $X$ and $S$ be a subset of $X$. The {\it product coefficient} parameter $a_S$  is the solution for the following system of equations
\begin{eqnarray}
	\mu(L(S)) &=& \frac{1}{2}(1 + a_s) \mu(S) \label{eqn:1}\\
	\mu(R(S)) &=& \frac{1}{2}(1 - a_s) \mu(S)  \label{eqn:2}
\end{eqnarray}	

%Dyadic Product Formula Representation
%$X$ with binary set system $B$ whose non-leaf sets are $B_n$ and $\mu=\mu(X) \prod_{S \in B_n} (1 + a_Sh_S)\, dy,$ where $a_S \in [-1,1]$ and $h_S$	is a Haar-like function

%%%
\begin{definition}
	Let $\mu$ be a dyadic measure on a dyadic set $X$ and $S$ be a subset of $X$. The {\it product coefficient} parameter $a_S$  is the solution for the following system of equations
	\begin{eqnarray}
		\mu(L(S)) &=& \frac{1}{2}(1 + a_s) \mu(S) \label{eqn:1}\\
		\mu(R(S)) &=& \frac{1}{2}(1 - a_s) \mu(S)  \label{eqn:2}
	\end{eqnarray}	
\end{definition}
A solution for \eqref{eqn:1}--\eqref{eqn:2} is unique if $\mu(S) \neq 0.$ If $\mu(S)=0$, we assign the zero value to the product coefficient, i.e., $a_S=0.$
Note that if $\mu(S)>0$ then solving  \eqref{eqn:1}--\eqref{eqn:2} for $a_s$ gives 
\begin{equation}
	a_s=\dfrac{\mu(L(S)) - \mu(R(S)) }{\mu(S)}
\end{equation}

%%%

The product coefficients are bounded, $|a_S| \leq 1.$  In what follows, we use a Haar-like function $h_S$ defined as 

\begin{equation}\label{Haar-like}
	1 \mbox{ on } L(S),\, -1 \mbox{ on } R(S), \mbox{ and } 0 \mbox{ on } X-S.
\end{equation}	

%$h_S=\left\{\begin{array}{l r}
	%1    & \mbox{ on }  L(S) \\
	%-1    & \mbox{ on }  R(S) \\
	% 0  & \mbox{otherwise}   \\
	%\end{array}\right.$
	
\begin{example}[Formula for a scale 0 dyadic measure]
	Let $X=[0,1]$ and let there be a non-negative measure $\mu$ such that $\mu(X)=1,\, \mu(L(X))=\frac{1}{4}$ and $\mu(R(X))=\frac{3}{4}.$ Let $a=a_X$ be the product coefficient which is the solution for the system of equations
	
	\begin{eqnarray}
		\mu(L(X)) &=& \frac{1}{2}(1 + a) \mu(X) \label{eqn:1X}\\
		\mu(R(X)) &=& \frac{1}{2}(1 - a) \mu(X)  \label{eqn:2X}.
	\end{eqnarray}	
	
	Subtracting \eqref{eqn:2X} from \eqref{eqn:1X} we obtain $a=\dfrac{\mu(L(X))-\mu(R(X))}{\mu(X)}=-\dfrac{1}{2}.$

\end{example}

Since, $dy(X)=1$ and $dy(L(X))=\frac{1}{2}=dy(R(X))$ then by the product formula form,
\begin{equation}
	\mu=\mu(X)(1+ah)dy,
\end{equation}
where $h$ is the Haar-like function with $S=X.$ 

In this paper, we use the counting measure on Lidar data points instead of Borel measure on intervals. We compute product coefficients in scale 0 (only one coefficient), scale 1 (two coefficients) and scale 2 (3 coefficients.)

The product formula for non-negative measures in $X=[0,1]$ using the product factors $a_S$ first appeared in \cite{Fefferman}. We present the representation lemma for dyadic sets extracted from \cite{Linda}.

%%%

\begin{lemma}[Dyadic Product Formula Representation]\label{lemma1}

	Let $X$ be a dyadic set with binary set system $B$ whose non-leaf sets are $B_n$.
	
	\begin{enumerate}
		\item A non-negative measure $\mu$ on $X$ has a unique product formula representation 
		\begin{equation}
			\mu=\mu(X) \prod_{S \in B_n} (1 + a_Sh_S)\, dy
		\end{equation}
		where $a_S \in [-1,1]$ and $a_S$ is the product coefficient for $S.$

		\item Any assignment of parameters $a_S$ for $(-1,1)$ and choice of $\mu(X)>0$ determines a measure $\mu$ which is positive on all sets $S$ on $B$ with product formula
		${\mu=\mu(X) \prod_{S \in B_n} (1 + a_Sh_S)\, dy}$
		whose product coefficients are the parameters $a_S.$
		
		\item Any assignment of parameters $a_S$ from $[-1,1]$ and choice of $\mu(X)>0$ determines a non-negative measure $\mu$ with product formula ${\mu=\mu(X) \prod_{S \in B_n} (1 + a_Sh_S)\, dy}.$ The parameters are the product coefficients if they satisfy the constraints:
		\begin{enumerate}
			
			\item If $a_S=1$, then the product coefficient for the tree rooted at $R(S)$ equals 0.
			\item If $a_S=-1$, then the product coefficient for the tree rooted at $L(S)$ equals 0.
		\end{enumerate}

	\end{enumerate}

\end{lemma}

\subsection{Real world applications}\label{subsec: applications}
 Advancement on this project will be of great
impact in studying climate change. One of the most important structural properties of vegetation (leaf area index) is related to the rate at which forests grow and sequester carbon which is also an important factor for micro climates within
forests that may help mitigate some of the initial impacts of climate change.

Differentiation of photosynthetic components (leaf, bushes or grasses) and non-photosynthetic components (branches or stems) by
3D terrestrial laser scanners (TLS) is of key importance to understanding the spatial distribution of the radiation
regime, photosynthetic processes, and carbon and water exchanges of the forest canopy. Research suggests that
woody canopy components are a major source of error in indirect leaf area index (LAI) estimates. We expect that
the use of a deep learning framework will improve the accuracy of the quantification of woody material, hence
improving the accuracy of LAI estimates (see \cite{Zheng2009} for background).

\subsection{Computing product coefficients on Lidar}\label{subsec: PCs on lidar}

The main purpose of this section is to show how to compute the product coefficients defined in Sec.\,\ref{PCs}

Applying the product formula representation as seen on Lemma\,\ref{lemma1} to a counting measure derived to our 3D point cloud LIDAR data. Recall that each point has been labeled as high vegetation, ground, building or low vegetation. 

Previous work \cite{BaIzMcNeSh:2012} had examined this same data using a multi-scale SVD approach to build a support vector machine (SVM) based classification rule that could, with high accuracy, reproduce the vegetation/ground labeling. In \cite{Linda} product coefficient parameters are used as features instead of multi-scale SVD parameters. Note that this work was done on only two classes. We are doing multiclass classification and adding the product coefficients computed locally, point by point. We use KNN and random forest classifiers to see the improvement of the classification after applying PCA to data with the new generated features. 
	
In \cite{BaIzMcNeSh:2012} the experiment showed that decision rules for distinguishing two measures (here ``vegetation and ``ground'') could be approximately inferred from histograms of the product coefficients. While the metrics were not as good as for multi-scale SVD, the method did provide a transparent rationale for the decision rule.

\begin{figure}\label{fig:binary}
\centering	
\includegraphics[scale=0.6]{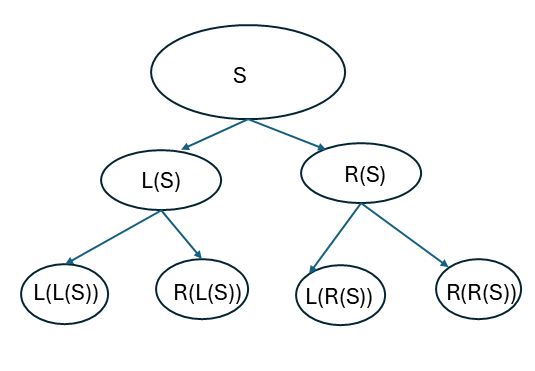}
\caption{The diagram represents the binary tree for a general set $S.$ The diagram includes the notation used in the computation of the seven product coefficients. There are $2^0, 2^1$ and $2^2$ product coefficients per level. }

\end{figure}

\begin{figure}
\includegraphics[scale=0.5]{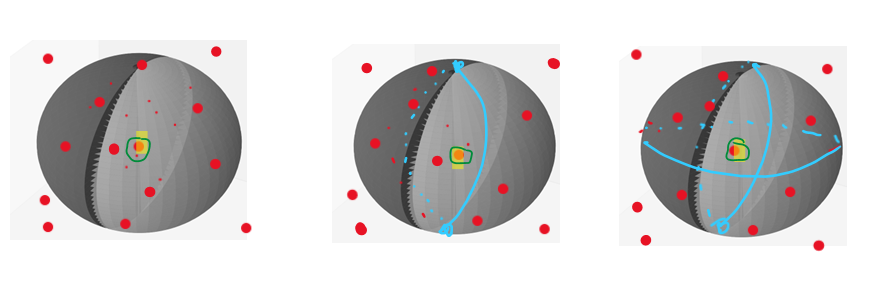}
\caption{When computing product coefficients per point $(x_i, y_i, z_i)$, consider a sphere $S+i$of radius 2. We first slice the sphere in along the $x$-axis and compute the first product coefficient $a_S$, then slice the sphere along the $y-$ axis and compute two product coefficients. One for the left child $L(S_i)$ and the other for the right child $R(S_i).$ Last, we slice along the $z-$ axis and compute the the last four product coefficients.} 
\label{fig:spheres}
\end{figure}

The general procedure to compute product coefficients goes as follow:

\begin{enumerate}
\item Normalize the original 3D point cloud to fit on the unit cube ${[0,1]}^3.$ Count the points of the Lidar dataset $S.$

\item For each fixed data point $(x_i,y_i,z_i)$, consider a sphere $S_i$ of radius 2.  Slice the sphere along the $x-$ axis direction and compute the product coefficient $a_S$ by counting the number of points in the left child $L(S)$ and the right child $R(S_i)$. The left child, $L(S_i)$ consists on all the points on the left section of the sphere and the right child the number of points in the right section of the sphere. 

\item Slice the sphere in the direction of the $y-$ axis and compute the product coefficients $a_{L(S_i)}$ and $a_{R(S_i)}.$ Notice that now the initial parent set is $L(S_i)$ or $R(S_i)$.

\item Finally, slice the sphere along the $z-$axis and compute the product coefficients $a_{L(L(S_i)},\, a_{R(L(S_i))},\, a_{L(R(S_i))}$ and $a_{R(R(_iS))}.$ See Fig,.\,\ref{fig:binary} for illustration of the binary tree and the notation for the children at each level. 
\end{enumerate}
This above procedure is shown in Fig.\,\ref{fig:spheres}. We have a total of seven product coefficients for each point $(x_i,y_i,z_i$).

We end up with seven new features and add them to the original spatial features $x,y$ and $z.$ 

\subsection{Classification experiments}
\label{subsec: PCA and PCs experients}

This section aims to examine the performance differences between two classification algorithms, K-Nearest Neighbors (KNN, with \( k=10 \)) and Random Forest (with 100 trees), in classifying the four LiDAR-derived classes in our original dataset. We compare this with classification based on a set of seven newly derived features obtained from two levels of product coefficients computed within a local neighborhood. This neighborhood is defined by a sphere of radius 2 centered at each data point, structured in a dyadic tree, and utilizing a counting measure with Principal Component Analysis (PCA) applied to retain up to three principal components.

The Nearest Neighbor classifier, particularly its K-Nearest Neighbors variant, is ideal for geospatial data as it utilizes spatial relationships between data points for classification and prediction. It determines the class or value of a new data point based on its proximity to neighboring points. On the other hand, Random Forest classifiers are highly effective for geospatial data analysis due to their ability to handle complex, high-dimensional datasets, capture non-linear relationships, and offer insights into feature importance.

Our proposed methodology offers several advantages. First, since dimensionality reduction is applied without using the data labels, this step minimizes the risk of overfitting. Second, while the original classifiers do not perform well on the base dataset, adding product coefficient features from localized neighborhoods and applying PCA with multiple principal components significantly enhances classification accuracy.

We conducted two main sets of experiments. In the first set, classification was performed using only the spatial coordinates (x, y, and z) as features, with both KNN and Random Forest classifiers. We then compared these results with classifications that included the seven new product coefficient features computed in the local neighborhood. In the second set, we applied the same classifiers using the generated seven features and implemented PCA to extract between three and ten principal components, which were used as inputs to the classifiers.

We use scikit learn in all experiments  \cite{pedregosa2018scikitlearnmachinelearningpython} and PyTorch for the autoencoder \cite{paszke2019pytorchimperativestylehighperformance}. The full source code is available at \url{https://github.com/rskarkare/Product-Coefficients}.
%We had little data for training, so we use a ``sliding window'' technique (see figure \ref{fig: sliding} for illustration) in order to produce more data points (each point has higher dimension depending on the window size) and improve the performance of the neural network. Overlapping windows and non-overlapping windows are considered. Note that the input data that gives the least number of data points corresponds to window size 1, which includes the original data together with all runs for a fixed scenario at one single time step. The windows don't overlap.

The main steps of our proposed algorithm, incorporating product coefficients as features along with Principal Component Analysis (PCA) and autoencoders, are as follows:

\begin{enumerate}
	\item \textbf{Feature Generation with Product Coefficients:} Starting with the original dataset, we generate a new feature set $X$ by computing product coefficients within localized neighborhoods. Specifically, we define a ball of radius 2 centered at each data point and use the points within this ball as a "parent" set for calculating the first-level product coefficient. In the subsequent level, two product coefficients are computed—one for the left child and one for the right child of this dyadic tree structure. At the following level, we compute $2^2$ additional product coefficients. This process yields seven new features for each data point. These features are then normalized to fit within the unit cube.
	
	\item \textbf{Dimensionality Reduction using PCA or autoencoder:} We apply PCA to the newly generated input data $X$ to create a transformed input layer $Z$ of varying dimensions, specifically with 3 to 10 principal components. This results in eight different experimental setups, each with a unique number of principal components.We do the same using the latent layer of an autoencoder. In this paper, we use the number of components of PCA as the dimension of the latent layer of the autoencoder. 
	
	\item \textbf{Classification with KNN, Random Forest} The transformed input $Z = XM \in \mathbb{R}^{n}$, where $M$ is the matrix of principal components derived from the covariance matrix of the input data, is then used as input to two classifiers—K-Nearest Neighbors (KNN) and Random Forest. Each experiment evaluates classifier performance as the dimensionality of $Z$ increases from $n=3$ to $n=10$. The cross-validated F1 scores for each setup are presented in Tables~\ref{tab:f1_macro_clean} and \ref{tab:f1_macro_knn}. Hyperparameters for RF and KNN are give on tables \ref{tab: hyper RF} and \ref{tab: hyper KNN}.
\end{enumerate}

\begin{table}[h]
	\centering
	
	\begin{tabular}{|l|l|p{8cm}|}
		\hline
		\textbf{Parameter} & \textbf{Value} & \textbf{Description} \\
		\hline
		\texttt{n\_estimators} & 200 & Number of trees in the forest. \\
		\texttt{random\_state} & 42 & Seed used by the random number generator (for reproducibility). \\
		\texttt{n\_jobs} & -1 & Number of jobs to run in parallel (uses all available processors). \\
		\texttt{class\_weight} & \texttt{balanced} & Adjusts weights inversely proportional to class frequencies in the input data. \\
		\hline
	\end{tabular}
	\caption{Random Forest Hyperparameters}
	\label{tab: hyper RF}
\end{table}

\begin{table}[h]
	\centering
	\begin{tabular}{|l|l|p{8cm}|}
		\hline
		\textbf{Parameter} & \textbf{Value} & \textbf{Description} \\
		\hline
		\texttt{n\_neighbors} & 10 & Number of neighbors to use for classification. \\
		\texttt{n\_jobs} & -1 & Number of parallel jobs to run (uses all available processors). \\
		\hline
	\end{tabular}
	\caption{K-Nearest Neighbors Hyperparameters}
	\label{tab: hyper KNN}
\end{table}

This structured approach allows us to evaluate the impact of localized product coefficients and PCA-based dimensionality reduction on classification accuracy.

%New table//
%\begin{table}[ht]
%	\centering
%	
%	\begin{tabular}{c c c c c}
	%		\hline
	%		\textbf{n\_features} & \textbf{AE Direct} & \textbf{AE $\rightarrow$ RF} & \textbf{Nystroem $\rightarrow$ RF} & \textbf{PCA $\rightarrow$ RF} \\
	%		\hline
	%		3  & 0.7788 & 0.8729 & 0.5123 & 0.5045 \\
	%		4  & 0.6798 & 0.8363 & 0.5275 & 0.5455 \\
	%		6  & 0.6579 & 0.7768 & 0.5069 & 0.5782 \\
	%		10 & 0.8002 & 0.7625 & 0.5751 & 0.5827 \\
	%		\hline
	%	\end{tabular}
%\caption{Combined F1-macro results for different feature extraction methods.}
%\label{tab:f1_macro}
%\end{table}
%We reduce the dimension of the space performing PCA. The cumulative variance is computed to get an estimate for the number of components. We end up choosing 20 components as the cumulative variance is more than $95\%.$
%
%After augmenting the dimension of the input layer, we perform  PCA to reduce the dimension to 20, then we train a neural network with two hidden layers ($L=4$). The input layer has dimension $d^{(0)}=20$ and the output, $d^{(3)}=100$, so the output layer contains $T$ the stopping times for each agent. The hidden layers have dimensions $d^{(1)}=50$ and $d^{(2)}=70.$  The time step used in all experiments is 0.05 seconds. 

\begin{table}[ht]
	\centering
	\begin{tabular}{c c c c c}
		\hline
		\textbf{number of PCs features} & \textbf{AE Direct} & \textbf{AE $\rightarrow$ RF} & \textbf{Nystroem $\rightarrow$ RF} & \textbf{PCA $\rightarrow$ RF} \\
		\hline
		3  & 0.7788 & 0.8729 & 0.5079 & 0.5045 \\
		4  & 0.6798 & 0.8363 & 0.5226 & 0.5455 \\
		6  & 0.6579 & 0.7768 & 0.5077 & 0.5782 \\
		10 & 0.8002 & 0.7625 & 0.5798 & 0.5827 \\
		\hline
	\end{tabular}
	\caption{Combined F1-macro results for different components.}
	\label{tab:f1_macro_clean}
\end{table}

\begin{table}[ht]
	\centering
	\begin{tabular}{c c c c c}
		\hline
		\textbf{components} & \textbf{AE Direct} & \textbf{AE $\rightarrow$ KNN} & \textbf{Nystroem $\rightarrow$ KNN} & \textbf{PCA $\rightarrow$ KNN} \\
		\hline
		3  & 0.8499 & 0.8787 & 0.5186 & 0.5035 \\
		4  & 0.8539 & 0.9172 & 0.5237 & 0.5383 \\
		5  & 0.8595 & 0.9304 & 0.5209 & 0.5590 \\
		6  & 0.8615 & 0.9414 & 0.5161 & 0.5623 \\
		7  & 0.8614 & 0.9428 & 0.5127 & 0.5751 \\
		8  & 0.8641 & 0.9498 & 0.5589 & 0.5730 \\
		9  & 0.8621 & 0.9538 & 0.5614 & 0.5792 \\
		10 & 0.8645 & 0.9502 & 0.5647 & 0.5741 \\
		\hline
	\end{tabular}
	\caption{Combined F1-macro results for AE, Nystroem, and PCA with KNN classifier. Hidden dimension layer for AE is the same number of components that we are using for PCA}
	\label{tab:f1_macro_knn}
\end{table}

Table \ref{tab:f1_macro_knn} reports the F1-macro performance obtained when maintaining the same total number of features and PCA components, with the aim of using PCA primarily as a covariance regularization method rather than a dimensionality reduction technique. In this configuration, the latent dimensionality of the Autoencoder (AE) is set equal to the number of PCA components, and the resulting feature representations are used as predictors for both K-Nearest Neighbors (KNN) and Random Forest (RF) classifiers. Notably, the AE - KNN framework exhibits consistently superior performance across all tested component sizes. For instance, it achieves an F1-macro of 0.8787 with three components and improves steadily to 0.9538 with nine components, slightly decreasing to 0.9502 at ten components. This pattern suggests that the nonlinear feature representations produced by the autoencoder effectively preserve class structure and enhance local neighborhood relationships, which are particularly advantageous for distance-based classifiers such as KNN. In contrast, the PCA → KNN results remain substantially lower, ranging from 0.5035 to 0.5792, indicating that linear projections are less effective in capturing the intrinsic geometry of the data manifold. The Nystroem - KNN method shows similar limitations, with scores fluctuating around 0.51–0.56. Overall, these results reinforce the benefit of combining autoencoder-derived latent features with nonparametric classifiers, demonstrating that the AE - KNN model provides a robust and discriminative representation that surpasses both kernel and linear alternatives under comparable feature conditions.

Table \ref{tab:f1_macro_clean} presents the F1-macro performance obtained using different dimensionality or covariance reduction methods such as PCA and the latent layer of an autoencoder (or encoder) as the number of product coefficients used as features increases. A general trend of improvement can be observed with the inclusion of additional product coefficients, indicating that richer feature representations contribute positively to classification performance. Among all evaluated approaches, the Autoencoder combined with the Random Forest classifier (AE - RF) consistently achieved the highest F1-macro values, with a peak performance of 0.8729 when using three components. Although slight variations occur as the dimensionality increases (e.g., 0.8363 for four components and 0.7768 for six components), the AE -RF configuration maintains superior performance compared to the Nystroem - RF and PCA - RF methods across all tested settings. The PCA - RF results, ranging from 0.5045 to 0.5827, remain notably lower, highlighting the limitations of linear transformations in capturing complex, nonlinear relationships within the data. In contrast, the autoencoder’s nonlinear latent representations enable the Random Forest to better discriminate between classes, resulting in improved overall generalization. These findings demonstrate that the proposed AE-based framework outperforms the PCA-based approach presented in previous studies, emphasizing the advantages of deep representation learning for enhanced classification of multidimensional data.

Table~	\ref{tab:autoencoder_params} shows all gthe hyperparamters used for the autoencoder. We have used PyTorch \cite{paszke2019pytorchimperativestylehighperformance}

\begin{figure}
	\includegraphics[scale=0.6]{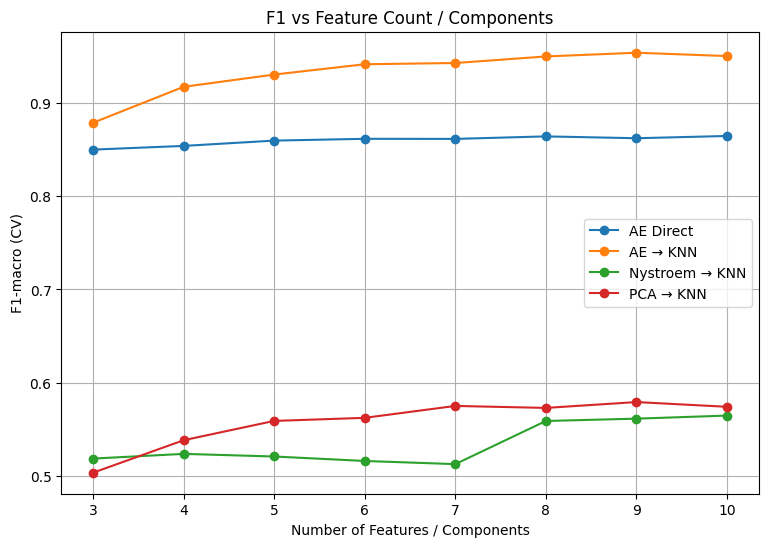}	
	\caption{F1 macro scores as number of features and components increase}
	\label{fig:F1plot}
\end{figure}

Figure~\ref{fig:F1plot} showcases the trends of the F1 scores.

\begin{table}[h!]
	\centering
	
	\begin{tabular}{ll}
		\hline
		\textbf{Parameter} & \textbf{Value} \\
		\hline
		Framework & PyTorch \\
		Hidden layers & 2 \\
		Hidden layer dimensions & 30, latent dimension $= n_{\text{features}}$ \\
		Activation functions & ReLU(), LeakyReLU() \\
		Batch size & 256 \\
		Number of epochs & 30 \\
		Optimizer & Adam \\
		Learning rate & $1 \times 10^{-3}$ \\
		\hline
	\end{tabular}
	\caption{PyTorch Autoencoder Training Parameters}
	\label{tab:autoencoder_params}
\end{table}

\section{Conclusion and Future Research Directions}
\label{summary}
In this paper, we demonstrated that enriching the original 3D LiDAR point cloud data with additional product coefficients (mathematical quantities originanted in the theoretical setting on the are of measure theory )computed on dyadic trees across local neighborhoods (sphere with radius 2) can improve classification performance. By combining these features with both linear and nonlinear dimensionality reduction techniques, we evaluated their impact using two well-established classifiers, Random Forest (RF) and K-Nearest Neighbors (KNN).

Our results indicate that while PCA remains effective as a covariance reduction method, its linear nature limits its capacity to capture the complex, nonlinear relationships inherent in geospatial data. In contrast, the Autoencoder (AE)-based frameworks, particularly AE - RF and AE - KNN, consistently outperform PCA and Nystroem methods under equivalent feature dimensions. The AE - KNN configuration achieved the best overall performance, reaching an F1-macro of 0.9538 with nine components, representing a significant improvement over the PCA-based results reported in our previous study. This improvement suggests that the nonlinear latent representations learned by the AE better preserve neighborhood structure and class separability, thereby enhancing the discriminative power of subsequent classifiers.

The dimensionality reduction techniques employed in this study also effectively reduced feature covariance, contributing to improved classifier generalization. Nonetheless, it is important to recognize that alternative nonlinear methods—such as variational autoencoders, diffusion maps, or manifold learning approaches—may yield comparable or superior results under different experimental conditions. Future work will investigate these alternatives and extend the analysis to larger datasets to assess scalability and computational efficiency.

Although computational cost was not addressed here due to the modest dataset size, future research will focus on efficiency and real-time applicability using larger-scale datasets, such as the Golden Gate Bridge LiDAR dataset comprising approximately 15 million points. Overall, this study underscores the value of nonlinear representation learning and feature enrichment through product coefficients as powerful tools for improving the accuracy of geospatial point cloud classification.\\

Future directions include: 
\begin{itemize}
	
	\item Extending the current framework to include variational autoencoders (VAEs), diffusion-based models, or graph neural networks (GNNs) to further explore nonlinear manifold structures within LiDAR point clouds.
	
	\item Evaluating the computational efficiency and scalability of the AE–KNN and AE–RF frameworks using larger and more complex datasets, such as the Golden Gate Bridge LiDAR dataset and other urban-scale geospatial collections.
	
%	\item Investigating the impact of feature normalization, data augmentation, and noise regularization on the stability and robustness of learned latent representations.
	
	\item Incorporating spatial–contextual features, such as local surface curvature, planarity, or intensity gradients, into the computation of product coefficients to better capture geometric relationships.
	
	\item Developing interpretable representations by analyzing latent-space clustering and feature importance to understand how autoencoder-derived features contribute to class separability.
	
	\item Comparing the proposed framework with additional covariance reduction and manifold learning methods, including t-SNE, UMAP, and kernel PCA, to quantify the benefits of deep nonlinear embedding over traditional techniques.
	
	\item Exploring hybrid learning strategies, such as semi-supervised or self-supervised training, to reduce dependence on labeled data while maintaining classification accuracy.
	
\end{itemize}

\begin{acknowledgement}
I want to thank Dr. Linda Ness for introducing me to product coefficients in Lidar and to Dr. Boris Iskra from AdTheorend Inc for helping us with debugging my code. 
\end{acknowledgement}

%\clearpage

%%% references %%%
\bibliographystyle{siam} 
\bibliography{references_lidar}
\end{document}